\documentclass[journal]{IEEEtran}
\usepackage{graphicx}
\usepackage{amsmath}
\usepackage{algorithm}
\usepackage{algpseudocode}
\usepackage{listings}
\usepackage{url}
\usepackage{cite}

\begin{document}
\title{FG-TreeSeg: Flow-Guided Tree Crown Segmentation without Instance Annotations}
\author{Pengyu Chen, Fangzheng Lyu, Sicheng Wang, and Cuizhen Wang%
\thanks{P. Chen, S. Wang, and C. Wang are with the Department of Geography, University of South Carolina, Columbia, SC 29208 USA (e-mail: pengyuc@email.sc.edu; SICHENGW@mailbox.sc.edu; CWANG@mailbox.sc.edu). F. Lyu is with the Department of Geography, Virginia Polytechnic Institute and State University, Blacksburg, VA 24061 USA (e-mail: fangzheng@vt.edu). Corresponding author: Fangzheng Lyu.}
}

\maketitle

\begin{abstract}
    Individual tree crown segmentation is an important task in remote sensing for forest biomass estimation and ecological monitoring. However, accurate delineation in dense, overlapping canopies remains a bottleneck. While supervised deep learning methods suffer from high annotation costs and limited generalization, emerging foundation models (e.g., Segment Anything Model) often lack domain knowledge, leading to under-segmentation in dense clusters. To bridge this gap, we propose FG-TreeSeg, a training-free framework for tree crown instance segmentation that transfers flow-based delineation from biomedical imaging to remote sensing. By modeling tree crowns as star-convex objects within a topological flow field using Cellpose-SAM, the FG-TreeSeg framework forces the separation of touching tree crown instances based on vector convergence. Experiments on the NEON and BAMFOREST datasets and visual inspection demonstrate that our framework generalizes robustly across diverse sensor types and canopy densities, which can offer a training-free solution for tree crown instance segmentation and labels generation.
\end{abstract}


\IEEEpeerreviewmaketitle

\section{Introduction}

\IEEEPARstart{T}{ree} crown instance segmentation remains a challenging problem in remote sensing due to the complex spatial structure of forest canopies \cite{9164904, 10767235}. Unlike discrete objects in general imagery, tree crowns in dense stands exhibit overlapping branches and ambiguous visual boundaries, making the definition of individual instances inherently difficult \cite{lucena2022combined}. Resolving these intricacies at scale is critical for large-scale tree inventories, where fine-grained, metropolis-wide canopy mapping has become essential for ecological management \cite{HE2022102667}.

Existing approaches to this problem face distinct methodological barriers. On one hand, supervised learning models suffer from a dependency on large-scale instance annotations. Generating such ground truth is labor-intensive and prone to subjectivity even among experts \cite{weinstein2019individual, lucena2022combined}, severely limiting model generalizability and scalability. On the other hand, recent foundation models such as the Segment Anything Model (SAM) \cite{kirillov2023segany} have significantly improved general-purpose segmentation. However, when applied to tree crown instance segmentation, these models often struggle in dense forests where adjacent crowns exhibit ambiguous boundaries \cite{teng_assessing_2025}. Some studies enhance instance separability by combining other data sources, such as tree height information from LiDAR \cite{straker_instance_2023}, Digital Surface Models (DSMs) \cite{teng2025bringingsamnewheights}, and multispectral features \cite{dersch2023towards}. Although effective, these multi-modal approaches often entail higher data acquisition costs and computational complexity, limiting their scalability.

To overcome these limitations without relying on auxiliary data or annotations, we draw inspiration from adjacent fields. We observe that semantic segmentation of tree canopies has reached a high level of maturity \cite{10342745, 11237054}, providing reliable foreground priors. Furthermore, tree crowns share star-convex morphological properties with biological cells. In biomedical imaging, flow-based methods such as CellViT \cite{horst2024cellvit} and Cellpose-SAM \cite{pachitariu_cellpose-sam_2025} successfully separate touching instances by predicting gradient fields, a mechanism highly transferable to canopy delineation. Tong and Zhang \cite{tong_individual_2025} established StarDist for supervised tree crown segmentation; here, we advance this geometric insight toward a training-free approach via SAM ViT cross-domain transfer.

Building on these insights, we propose FG-TreeSeg (Flow-Guided Tree Segmentation), a pipeline that enables training-free instance delineation by transferring flow-based dynamics from biomedical imaging to canopy analysis.

\section{Methodology}

The core methodological contribution of this study lies in domain-adaptive transfer: we reconceptualize tree crown delineation as a topological flow-convergence problem rather than a traditional boundary detection task. Our approach leverages the structural isomorphism between biological cells and tree crowns, both of which are star-convex objects, to transfer gradient flow dynamics from biomedical imaging to remote sensing for separating dense, touching instances without supervision. Crucially, we employ a semantic prior (SegFormer) to spatially regularize Cellpose-SAM, preventing background over-segmentation and enabling robust generalization.

\begin{figure}[ht]
    \centering
    \includegraphics[width=1\linewidth]{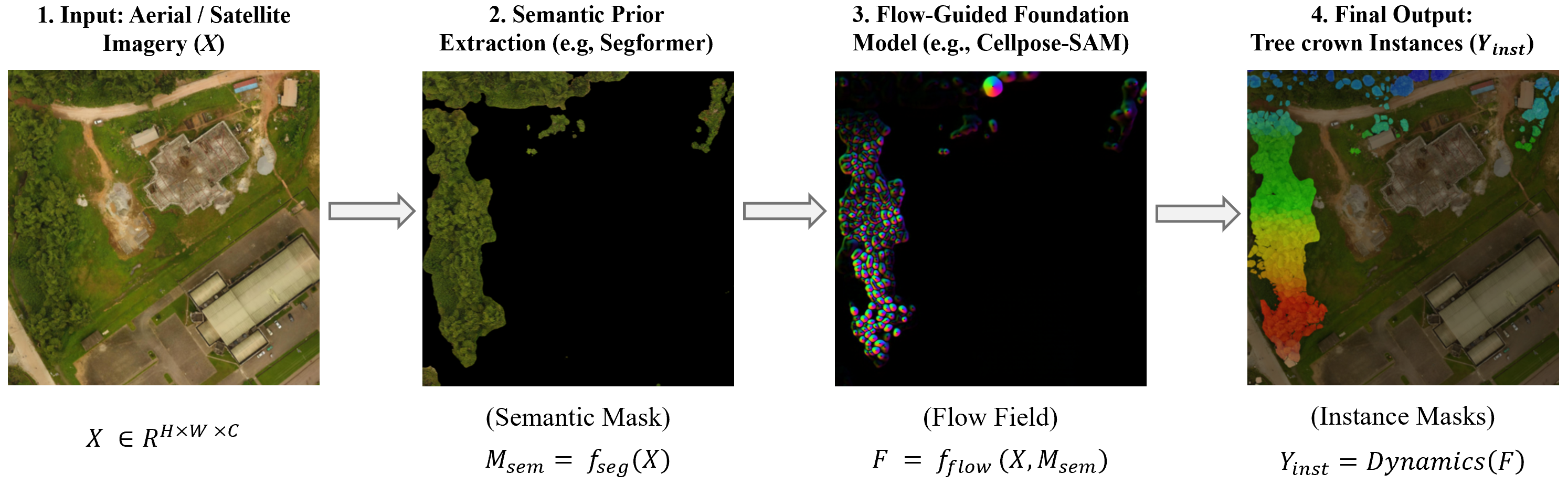}
    \caption{Overview of the training tree Tree Crown Segmentation Framework. The pipeline integrates a semantic prior to extract the canopy mask, followed by a cell segmentation model (Cellpose-SAM) that predicts gradient vector fields to isolate individual tree instances.}
    \label{fig:workflow}
\end{figure}

The workflow is illustrated in Figure \ref{fig:workflow}, executes this concept in two stages: 1) Canopy Semantic Constraint, which isolates the search space to valid canopy regions; and 2) Flow-guided Instance Segmentation, which forces the separation of individual crowns. The specific implementation details are described in Sections 2A and 2B, respectively.

\subsection{Semantic Segmentation}

In the FG-TreeSeg's first stage, a SegFormer model (MiT-B5 backbone)  trained on the OAM-TCD (OpenAerialMap-tree crown delineation) dataset \cite{veitch-michaelis_oam-tcd_2024} is used to perform binary semantic segmentation of tree canopies. The model achieves an $F_1$ score of 0.914 and an Intersection-over-Union (IoU) of 0.887, producing a semantic mask $M_{\text{sem}} \in {0,1}^{H \times W}$ that delineates canopy pixels (class 1) from background elements (class 0), including soil, roads, and built-up areas\cite{veitch-michaelis_oam-tcd_2024}.

This semantic mask serves as a spatial and semantic constraint for the subsequent instance segmentation stage. Although foundation models such as SAM and Cellpose-SAM exhibit strong generalization capabilities, their predictions remain sensitive to background textures in high-resolution aerial imagery \cite{teng_assessing_2025}. In particular, Cellpose-SAM was derived from SAM and further fine-tuned using large-scale cellular imagery \cite{pachitariu_cellpose-sam_2025}, which tends to generate spurious instance detections in visually complex non-canopy regions when applied without prior spatial filtering.

\begin{figure}[ht]
\centering
\includegraphics[width=0.8\linewidth]{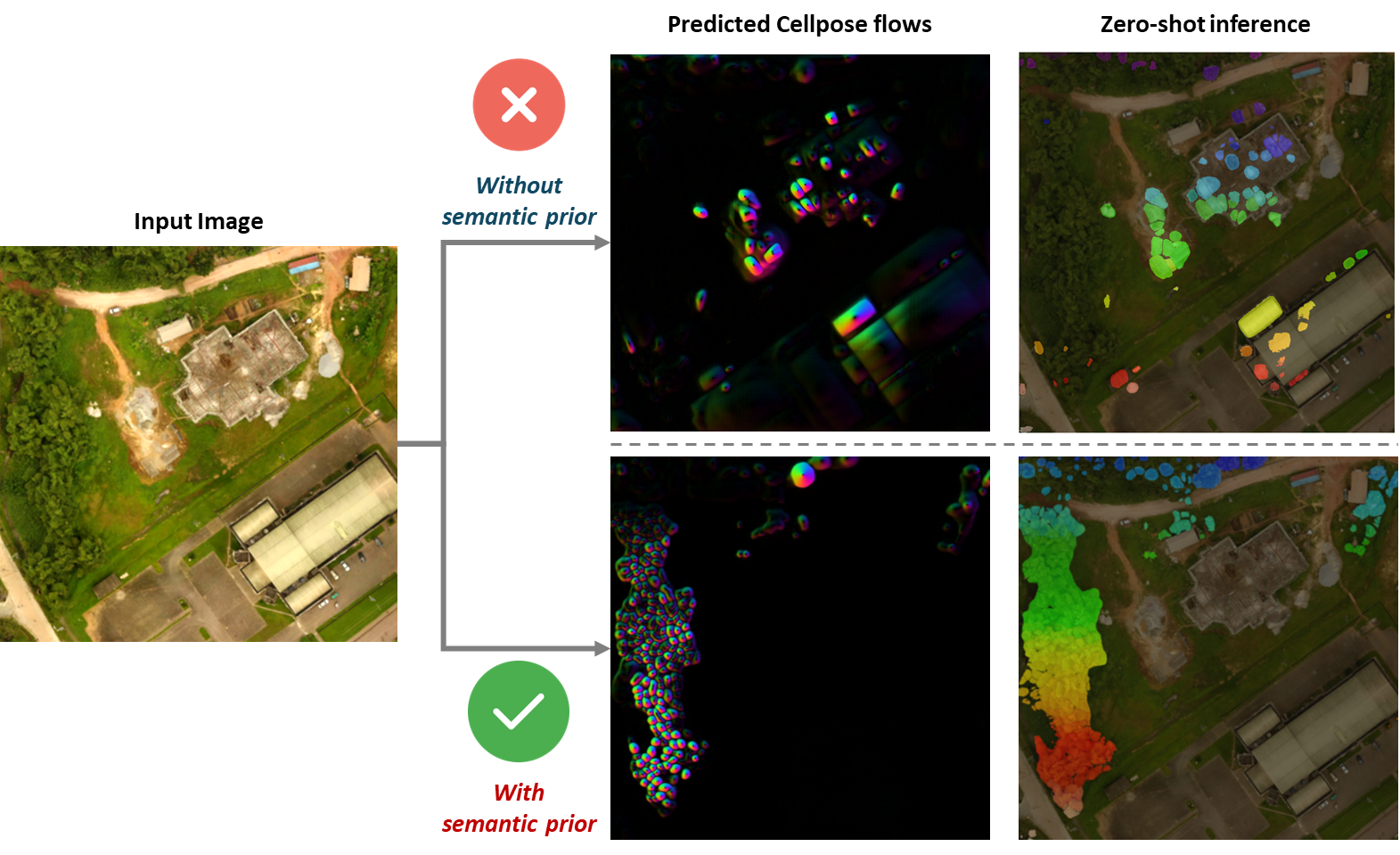}
\caption{Effect of semantic masking on instance segmentation.}
\label{fig:compare_semantic}
\end{figure}

As shown in Fig. \ref{fig:compare_semantic}, omitting semantic filtering results in pronounced over-segmentation over background surfaces such as grass and rooftops. By contrast, applying the semantic prior $M_{\text{sem}}$ effectively restricts instance inference to canopy regions, substantially reducing false positives and improving the stability of predictions.

Semantic segmentation is treated as a preprocessing step rather than a core contribution of this work. Given the maturity of canopy segmentation in remote sensing and the availability of high-performing pretrained models, SegFormer is employed as an off-the-shelf component to provide a reliable region of interest (ROI) for downstream instance segmentation.

\subsection{Flow-guided Instance Segmentation Framework}

\begin{figure}[ht]
    \centering
    \includegraphics[width=1\linewidth]{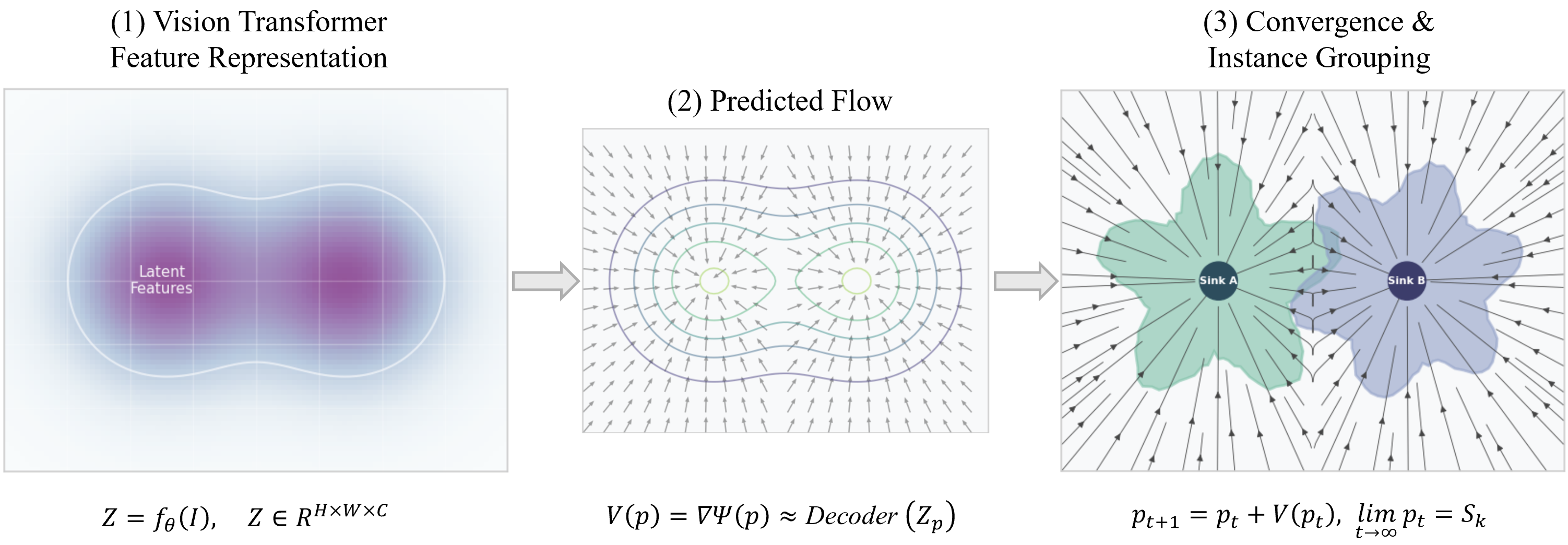}
    \caption{Conceptual framework of the proposed flow-based segmentation. (1) Feature Encoding: A ViT backbone extracts a latent representation $\mathbf{Z}$ capturing global shape context. (2) Flow Prediction: The network predicts a unit vector field $\mathbf{V}$ (gradient of a diffusion potential) that directs pixels toward object interiors. (3) Grouping: Segmentation is achieved by iteratively tracking pixel trajectories until they converge to stable sinks.}
    \label{fig:conceptualimage}
\end{figure}

FG-TreeSeg framework adapts the Cellpose-SAM architecture for tree-crown delineation by leveraging the geometric property that biological cells and tree crowns share a quasi-star-convex structure \cite{weigert2022, tong_individual_2025}. This inductive bias is particularly effective for separating dense, touching crowns in aerial imagery where explicit boundary cues are often ambiguous. As illustrated in Fig.~\ref{fig:conceptualimage}, the segmentation process is formulated as a dynamical system involving latent feature encoding, gradient field prediction, and iterative pixel convergence.

First, to capture high-level semantic context, the input image $\mathbf{I}$ is mapped into a shape-aware latent space using the SAM-based Vision Transformer (ViT) backbone. Unlike convolutional encoders with limited receptive fields, the ViT encoder, denoted as $\Phi_{\text{ViT}}$, aggregates global contextual information to produce a robust feature representation $\mathbf{Z}$:
\begin{equation}
    \mathbf{Z} = \Phi_{\text{ViT}}(\mathbf{I}).
\end{equation}
This representation implicitly encodes object interior structures and spatial dependencies, providing a rich basis for distinguishing individual instances within crowded canopies.

Subsequently, the network predicts a pixel-wise vector field $\mathbf{V}$ based on the local latent features $\mathbf{Z}_p$. Theoretically, this field represents the spatial gradient of an instance-specific potential surface $\Psi$, which exhibits local minima at object centers. The predicted flow vector at pixel $p$ is obtained by decoding the feature representation:
\begin{equation}
    \mathbf{V}(p) = \nabla \Psi(p) \approx \text{Decoder}(\mathbf{Z}_p),
\end{equation}
where $\text{Decoder}(\cdot)$ denotes the prediction head mapping latent features to 2D flow vectors. Geometrically, $\mathbf{V}(p)$ guides pixels inward toward the centroid (sink) of their respective instances. By approximating this gradient field, the network naturally handles topology; the flow diverges at the boundaries between touching crowns and converges within crown interiors, effectively converting segmentation into a flow prediction task.

Finally, instance masks are recovered through a flow-convergence grouping process during inference. Pixels are treated as particles in a dynamical system and are iteratively advected along the predicted stream lines using Euler integration:
\begin{equation}
    p_{\tau+1} = p_{\tau} + \mathbf{V}(p_{\tau}),
\end{equation}
where $\tau$ denotes the iteration step. As $\tau \to \infty$, pixels belonging to the same crown converge to a common stable fixed point $S_k$. The set of all pixels whose trajectories terminate at $S_k$ constitutes the $k$-th instance mask $\mathcal{M}_k$. This mechanism ensures robust instance separation solely through flow dynamics, eliminating the need for post-processing steps like watershed transforms or non-maximum suppression.

Algorithm~\ref{alg:inference} details this inference workflow, where the network outputs spatial flow gradients ($\Delta y, \Delta x$) directed at crown centers and a foreground probability map ($P$). During grouping, we set \textit{cellprob\_threshold = 0} to bypass redundant filtering, since the SegFormer mask already strictly defines the ROI. Additionally, we apply a relaxed \textit{flow\_threshold = 1} to ensure asymmetric, geometrically complex canopies are retained.

\begin{algorithm}[ht]
\caption{Inference Workflow for Tree Crowns}
\label{alg:inference}
\begin{algorithmic}[1]
\Require Input image $X$, semantic canopy mask $M_{\text{sem}}$
\State \textbf{Stage 1: Semantic masking}
\State $X_{\text{masked}} \leftarrow X \odot M_{\text{sem}}$

\State \textbf{Stage 2: Flow prediction}
\State $(\Delta y, \Delta x, P) \leftarrow \text{CellposeSAM}(X_{\text{masked}})$

\State \textbf{Stage 3: Flow-based grouping (gradient tracking)}
\State \textit{// The cell probability threshold is set to 0 to avoid filtering out candidate targets}

\State $Y_{\text{inst}} \leftarrow \text{AdvectAndCluster}(\Delta y, \Delta x, P; flow\_thresh=1, cellprob\_thresh=0)$

\State \Return $Y_{\text{inst}}$
\end{algorithmic}
\end{algorithm}

\section{Results}

This section evaluates the performance of the FG-TreeSeg framework. The analysis is organized into three components: a qualitative assessment of the model's generalization across varying scales and densities, a quantitative benchmark on the NEON and BAMFORESTS datasets, and a detailed visual inspection comparing the framework against supervised baselines in complex environments.

\subsection{Qualitative Assessment}

\begin{figure}[ht] 
\centering \includegraphics[width=0.9\linewidth]{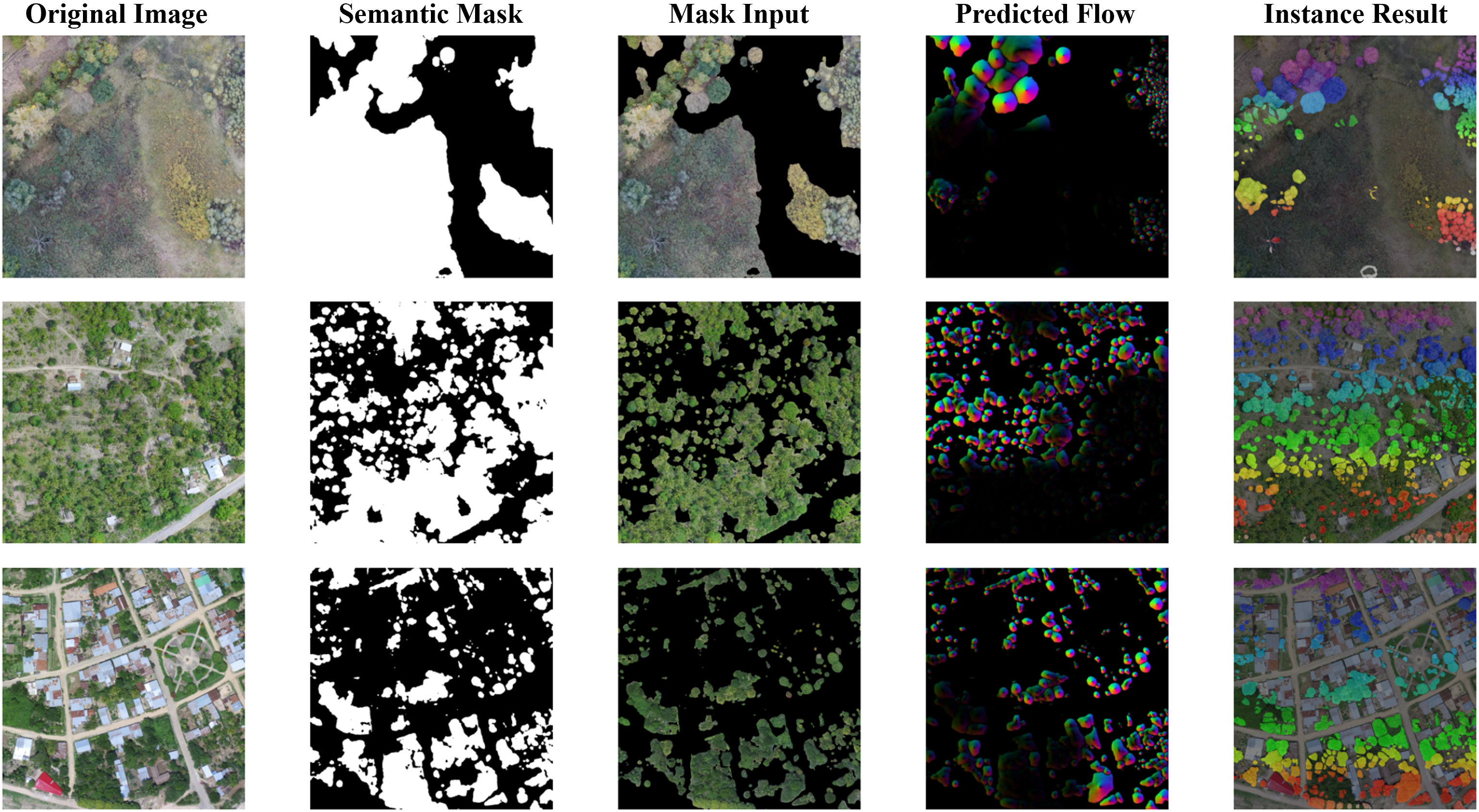} 
\caption{Results of the instance segmentation.} 
\label{fig:all_result} 
\end{figure}

Our experiments demonstrate that the FG-TreeSeg framework achieves stable accuracy and strong generalization across diverse land cover contexts without requiring any task-specific fine-tuning. As illustrated in Figure \ref{fig:all_result}, the model effectively handles the multi-scale nature of forestry. It exhibits a robust capability to delineate tree crowns of varying sizes, successfully capturing both large, isolated heritage trees and smaller, clustered saplings within the same inference pass.

\begin{figure}[ht] 
\centering \includegraphics[width=0.9\linewidth]{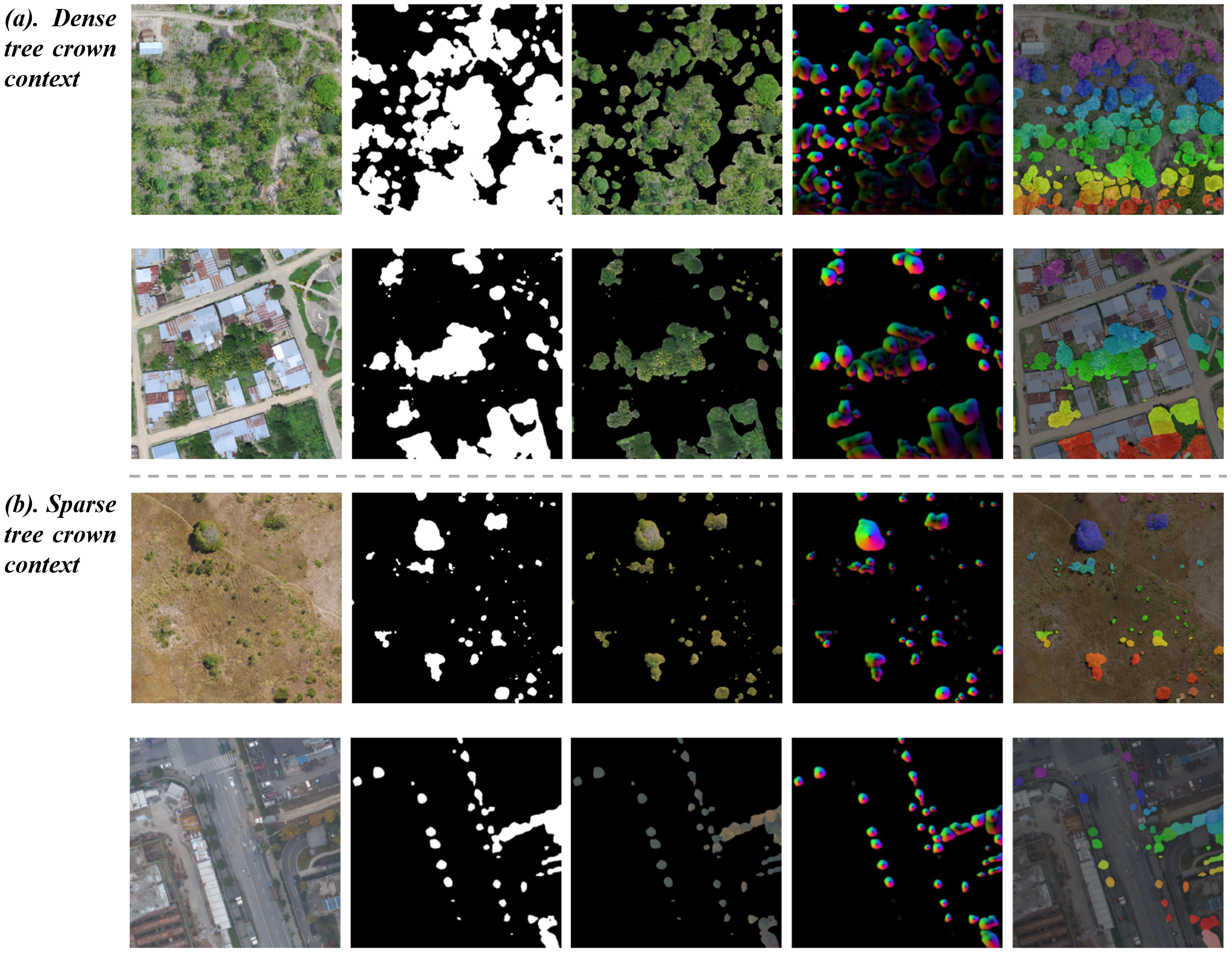} 
\caption{Performance comparison across varying canopy densities. While the model achieves near-perfect segmentation in sparse urban contexts (top), it also maintains high separability in dense, overlapping canopy environments (bottom) by leveraging flow field dynamics.} 
\label{fig:tree_dense_sparse} 
\end{figure}

We further analyzed the model's behavior under varying canopy densities to understand its operational boundaries. Figure \ref{fig:tree_dense_sparse} presents a comparative visualization between sparse urban scenarios and dense forest environments.

In sparse contexts, such as street trees or suburban parks, the approach performs exceptionally well. The semantic prior effectively isolates the Region of Interest (ROI), and the distinct spatial separation between instances makes the gradient flow prediction highly reliable.

In dense contexts, where tree crowns are heavily interconnected or overlapping, the flow-guided mechanism proves critical. Unlike traditional boundary-based methods that often under-segment fused canopies, our method successfully partitions dense clusters by identifying distinct topological centroids. However, we acknowledge that in extreme cases of canopy fusion, over-segmentation or boundary ambiguity persist. This suggests that while our approach is strong, these complex areas could still benefit from minimal manual refinement.

\subsection{Quantitative Performance}

To assess the generalizability of the FG-TreeSeg framework, we conducted a benchmark evaluation across two diverse datasets: NEON \cite{ben_weinstein_2022_5914554} (aerial RGB) and BAMFORESTS \cite{BAMFORESTS} (UAV VHR). The quantitative results are summarized in Table~\ref{tab:combined_benchmark}.

On the NEON dataset (Panel A), FG-TreeSeg achieves an mAP@50 of 42.30\%. While the supervised detection specialist DeepForest achieves higher accuracy (49.89\%) due to its optimized bounding-box regression, our training-free segmentation approach outperforms the point-supervised TreePseCo (41.68\%) \cite{lungo_vaschetti_treepseco_2025}. This demonstrates superior capability in separating touching crowns without explicit instance-level training.

On the BAMFORESTS dataset (Panel B), our framework achieves an mAP@50 of 67.31\%. Although its performance still lags behind supervised models like Mask R-CNN (69.05\%), it has achieved a level that is less than 2\% different from deeply trained professional models without the need for manual annotation or extensive training time. It confirms that the geometric priors leveraged by FG-TreeSeg effectively bridge the gap between annotation-free inference and supervised accuracy.

\begin{table}[ht]
\centering
\caption{Quantitative comparison on NEON and BAMFORESTS datasets.}
\label{tab:combined_benchmark}
\begin{tabular}{l c c c}
\hline
\textbf{Method} & \textbf{Source} & \textbf{Type} & \textbf{mAP@50} \\
\hline
\multicolumn{4}{l}{\textit{\textbf{Panel A: NEON Dataset (Aerial RGB)}}} \\
DeepForest & \cite{lungo_vaschetti_treepseco_2025} & Supervised (Box) & \textbf{49.89} \\
\textbf{FG-TreeSeg} &  & \textbf{Training-free} & 42.30 \\
TreePseCo & \cite{lungo_vaschetti_treepseco_2025} & Supervised (Point) & 41.68 \\
\hline
\multicolumn{4}{l}{\textit{\textbf{Panel B: BAMFORESTS Dataset (UAV VHR)}}} \\
Mask R-CNN & \cite{ruschhaupt_comparing_2025} & Supervised & \textbf{69.05} \\
Mask2Former & \cite{ruschhaupt_comparing_2025} & Supervised & 68.89 \\
\textbf{FG-TreeSeg} &  & \textbf{Training-free} & 67.31 \\
\hline
\end{tabular}
\end{table}
\subsection{Visual Inspection}

To validate the efficacy of our hybrid framework beyond standard metrics, we performed a visual comparison against representative baselines. We included the TCD dataset \cite{veitch-michaelis_oam-tcd_2024} in this visual analysis. Although TCD was excluded from the quantitative benchmark due to significant granularity mismatches, where ground truth annotations frequently aggregate dense clusters into single polygons, it serves as a critical stress test for evaluating model behavior in both dense and sparse canopy environments.

\begin{figure}[ht]
\centering 
\includegraphics[width=0.9\linewidth]{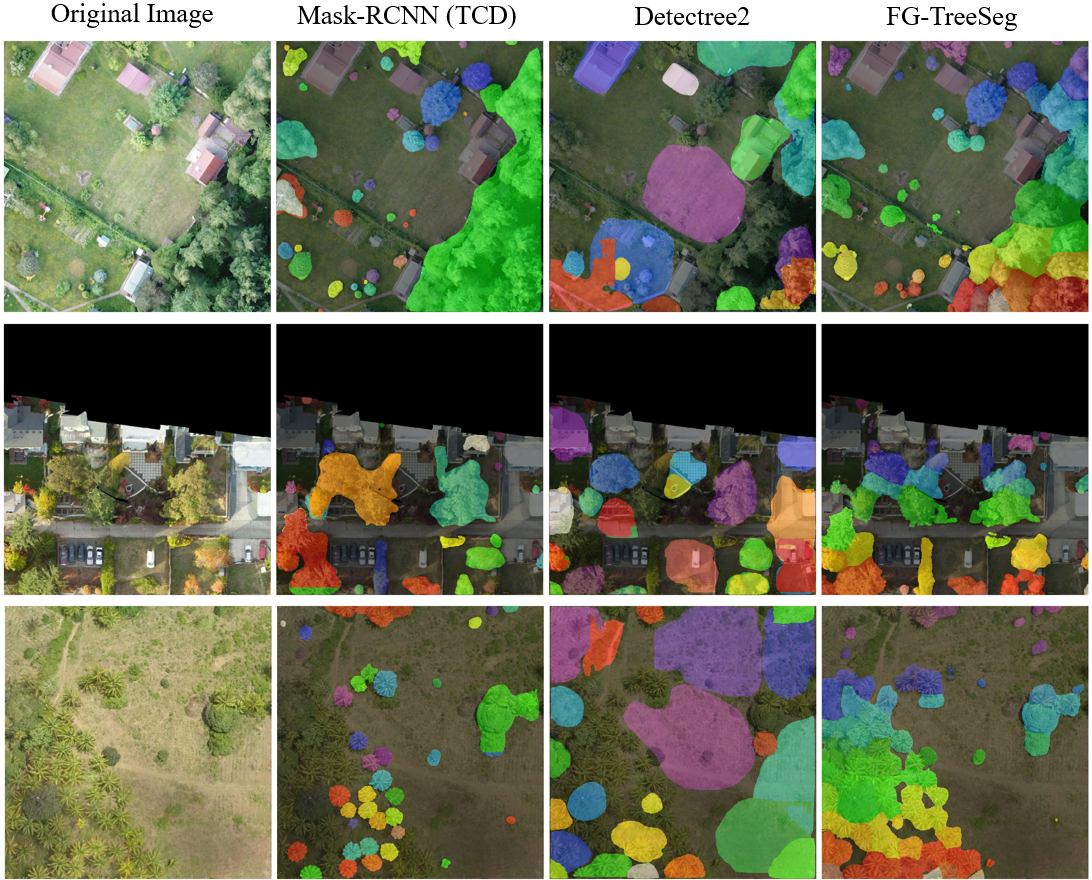} 
\caption{Comparison with Mask R-CNN (TCD) and Detectree2.} 
\label{fig:comparison} 
\end{figure}

The baselines exhibit failure modes on TCD's test dataset (Figure \ref{fig:comparison}). The supervised Mask R-CNN (trained on TCD) leans toward under-segmentation, frequently merging adjacent trees into single coarse clusters, reflecting the aggregation bias of its training data. Conversely, the domain-specific tool Detectree2 \cite{ball_accurate_2023} suffers from diminished adaptability, demonstrating low precision by classifying non-tree background features as crowns. In contrast, FG-TreeSeg achieves the optimal trade-off. The semantic prior acts as a robust filter to eliminate background noise, while the geometric flow mechanism enforces the topological separation of touching crowns. This resolves the aggregation issues common to standard supervised models, even in cases where the ground truth itself lacks fine-grained separability.

\begin{figure}[ht]
\centering 
\includegraphics[width=0.9\linewidth]{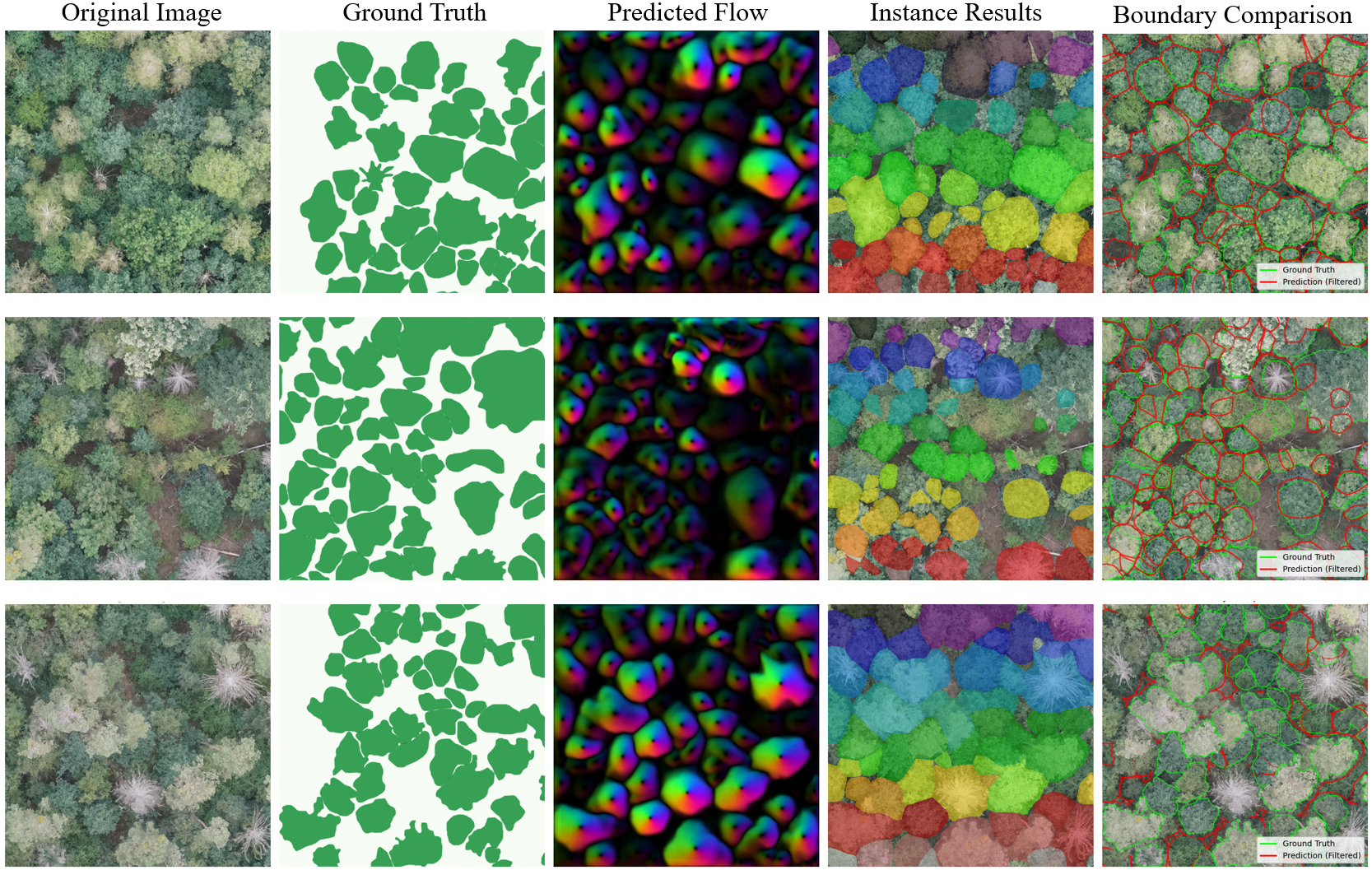}
\caption{Generalization test on the BAMFORESTS dataset.} 
\label{fig:BAMFOREST} 
\end{figure}

We also visually verified the model’s domain transfer capability on the BAMFORESTS dataset (Figure \ref{fig:BAMFOREST}). The resulting segmentations demonstrate that FG-TreeSeg reliably delineates accurate and well-defined boundaries in previously unseen domains, and does so without exhibiting the sensitivity and instability that are frequently reported for other foundation models (e.g., SAM \cite{teng_assessing_2025}).

\section{Discussion}

Unlike standard SAM variants that rely on explicit prompting and struggle with overlapping boundaries in automatic modes, FG-TreeSeg utilizes Cellpose-SAM's gradient flow fields to mathematically separate touching star-convex crowns. This fundamental difference prompted us to benchmark against fully supervised domain specialists rather than baseline SAMs for a more rigorous evaluation.

The qualitative discrepancy observed in the TCD evaluation points to a fundamental conflict in benchmarking: granularity mismatch. While FG-TreeSeg enforces the topological separation of distinct tree centers, datasets like TCD frequently aggregate dense clusters into single polygons. Valid fine-grained predictions would be paradoxically penalized as false positives, rendering standard quantitative evaluation unreliable. For this reason, we excluded TCD from quantitative benchmarks and retained it exclusively for visual analysis. This visual comparison explicitly demonstrates a critical nuance: while TCD-trained supervised models perform well on isolated single crowns, they severely under-segment dense canopies due to inherited annotation biases. In contrast, FG-TreeSeg robustly separates both isolated trees and overlapping clusters, visually proving its superior fine-grained separability despite the flawed ground truth.

\begin{figure}
    \centering
    \includegraphics[width=0.8\linewidth]{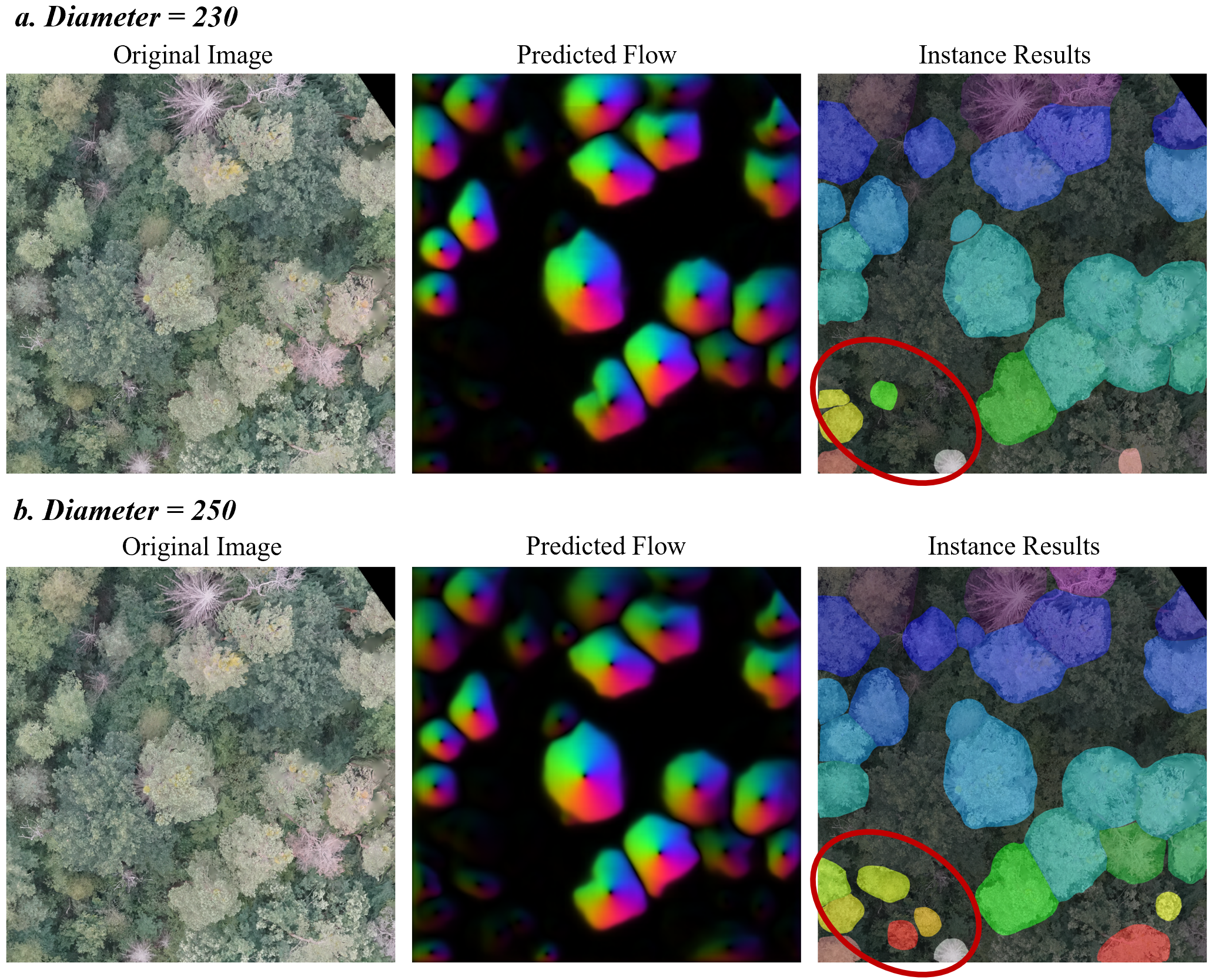}
    \caption{The impact of the diameter parameter on segmentation granularity. (a) Diameter = 230 pixels (small/young crowns); (b) Diameter = 250 pixels (mature stands). This flow-field convergence scale is readily calibrated via local species knowledge or brief visual inspection.}
    \label{fig:diameter_change}
\end{figure}

Furthermore, our framework incorporates an explicit parameter: the average crown diameter. As illustrated in Fig. \ref{fig:diameter_change}, this parameter governs the convergence scale of the flow field, where a smaller value encourages the separation of dense instances and a larger value promotes the aggregation of continuous canopies. We acknowledge that this parameter exhibits context-dependent sensitivity to stand density, species composition, and sensor resolution. Systematic calibration across diverse biomes and the transition towards a fully automated, annotation-free foundation model remain as critical future work.

\section{Conclusion}

In this letter, we present a training-free framework for individual tree crown segmentation that decouples Semantic Prior Extraction (via SegFormer) from Flow-Guided Instance Separation (via Cellpose-SAM). By leveraging geometric flow dynamics, our approach achieves robust annotation-free generalization, performing competitively with supervised methods on the benchmark NEON and BAMFORESTS datasets. 

While current absolute performance across the field remains below the threshold required for precise ecological accounting, FG-TreeSeg provides a scalable, out-of-the-box solution. Crucially, rather than replacing supervised learning, it offers a rapid method for generating preliminary instance masks. This accelerates annotation by shifting the effort from manual polygon drawing to boundary refinement, thereby supporting the efficient construction of massive datasets needed to train future forestry foundation models.

\bibliographystyle{IEEEtran}
\bibliography{references}

@article{kirillov2023segany,
  title={Segment Anything},
  author={Kirillov, Alexander and Mintun, Eric and Ravi, Nikhila and Mao, Hanzi and Rolland, Chloe and Gustafson, Laura and Xiao, Tete and Whitehead, Spencer and Berg, Alexander C. and Lo, Wan-Yen and Doll{\'a}r, Piotr and Girshick, Ross},
  journal={arXiv:2304.02643},
  year={2023}
}

@ARTICLE{9164904,
  author={Tusa, Eduardo and Monnet, Jean-Matthieu and Barré, Jean-Baptiste and Mura, Mauro Dalla and Dalponte, Michele and Chanussot, Jocelyn},
  journal={IEEE Geoscience and Remote Sensing Letters}, 
  title={Individual Tree Segmentation Based on Mean Shift and Crown Shape Model for Temperate Forest}, 
  year={2021},
  volume={18},
  number={12},
  pages={2052-2056},
  doi={10.1109/LGRS.2020.3012718}}

@ARTICLE{10767235,
  author={Ding, Ze and Zhang, Huaiqing and Wang, RuiSheng and Zhang, Li and Jiang, Hanxiao and Yun, Ting},
  journal={IEEE Geoscience and Remote Sensing Letters}, 
  title={A Dual-Branch Deep Learning Framework at the Grid Scale for Individual Tree Segmentation}, 
  year={2025},
  volume={22},
  number={},
  pages={1-5},
  doi={10.1109/LGRS.2024.3506223}}

@article{lucena2022combined,
  title={The combined use of UAV-based RGB and DEM images for the detection and delineation of orange tree crowns with Mask R-CNN: An approach of labeling and unified framework},
  author={Lucena, Felipe and Breunig, Fabio Marcelo and Kux, Hermann},
  journal={Future Internet},
  volume={14},
  number={10},
  pages={275},
  year={2022},
  publisher={MDPI}
}

@article{weinstein2019individual,
  title={Individual tree-crown detection in RGB imagery using semi-supervised deep learning neural networks},
  author={Weinstein, Ben G and Marconi, Sergio and Bohlman, Stephanie and Zare, Alina and White, Ethan},
  journal={Remote Sensing},
  volume={11},
  number={11},
  pages={1309},
  year={2019},
  publisher={MDPI}
}

@misc{teng2025bringingsamnewheights,
      title={Bringing SAM to new heights: Leveraging elevation data for tree crown segmentation from drone imagery}, 
      author={Mélisande Teng and Arthur Ouaknine and Etienne Laliberté and Yoshua Bengio and David Rolnick and Hugo Larochelle},
      year={2025},
      eprint={2506.04970},
      archivePrefix={arXiv},
      primaryClass={cs.CV},
      url={https://arxiv.org/abs/2506.04970}, 
}

@ARTICLE{10342745,
  author={He, Haiqing and Zhou, Fuyang and Xia, Yuanping and Chen, Min and Chen, Ting},
  journal={IEEE Journal of Selected Topics in Applied Earth Observations and Remote Sensing}, 
  title={Parallel Fusion Neural Network Considering Local and Global Semantic Information for Citrus Tree Canopy Segmentation}, 
  year={2024},
  volume={17},
  number={},
  pages={1535-1549},
  }

@inproceedings{weigert2022,
  author    = {Martin Weigert and Uwe Schmidt},
  title     = {Nuclei Instance Segmentation and Classification in Histopathology Images with Stardist},
  booktitle = {The IEEE International Symposium on Biomedical Imaging Challenges (ISBIC)},
  year = {2022},
  doi       = {10.1109/ISBIC56247.2022.9854534}
}

@article{straker_instance_2023,
	title = {Instance segmentation of individual tree crowns with {YOLOv}5: A comparison of approaches using the {ForInstance} benchmark {LiDAR} dataset},
	volume = {9},
	issn = {2667-3932},
	url = {https://www.sciencedirect.com/science/article/pii/S2667393223000169},
	doi = {https://doi.org/10.1016/j.ophoto.2023.100045},
	pages = {100045},
	journal = {{ISPRS} Open Journal of Photogrammetry and Remote Sensing},
	author = {Straker, Adrian and Puliti, Stefano and Breidenbach, Johannes and Kleinn, Christoph and Pearse, Grant and Astrup, Rasmus and Magdon, Paul},
	year = {2023},
}

@article{dersch2023towards,
  title={Towards complete tree crown delineation by instance segmentation with Mask R--CNN and DETR using UAV-based multispectral imagery and lidar data},
  author={Dersch, Sebastian and Schoettl, Alfred and Krzystek, Peter and Heurich, Marco},
  journal={ISPRS Open Journal of Photogrammetry and Remote Sensing},
  volume={8},
  pages={100037},
  year={2023},
  publisher={Elsevier}
}

@ARTICLE{11237054,
  author={Gui, Yuanyuan and Li, Wei and Wang, Yinjian and Xia, Xiang-Gen and Marty, Mauro and Ginzler, Christian and Wang, Zuyuan},
  journal={IEEE Journal of Selected Topics in Applied Earth Observations and Remote Sensing}, 
  title={Multimodal Uncertainty Robust Tree Cover Segmentation for High-Resolution Remote Sensing Images}, 
  year={2026},
  volume={19},
  number={},
  pages={114-128},
  doi={10.1109/JSTARS.2025.3631272}}

@article{pachitariu_cellpose-sam_2025,
	title = {Cellpose-{SAM}: superhuman generalization for cellular segmentation},
	url = {https://www.biorxiv.org/content/early/2025/05/01/2025.04.28.651001},
	doi = {10.1101/2025.04.28.651001},
	journal = {{bioRxiv}},
	publisher = {Cold Spring Harbor Laboratory},
	author = {Pachitariu, Marius and Rariden, Michael and Stringer, Carsen},
	year = {2025},
}

@Article{BAMFORESTS,
AUTHOR = {Troles, Jonas and Schmid, Ute and Fan, Wen and Tian, Jiaojiao},
TITLE = {BAMFORESTS: Bamberg Benchmark Forest Dataset of Individual Tree Crowns in Very-High-Resolution UAV Images},
JOURNAL = {Remote Sensing},
VOLUME = {16},
YEAR = {2024},
NUMBER = {11},
ARTICLE-NUMBER = {1935},
URL = {https://www.mdpi.com/2072-4292/16/11/1935},
ISSN = {2072-4292},
DOI = {10.3390/rs16111935}
}

@article{horst2024cellvit,
  title={Cellvit: Vision transformers for precise cell segmentation and classification},
  author={H{\"o}rst, Fabian and Rempe, Moritz and Heine, Lukas and Seibold, Constantin and Keyl, Julius and Baldini, Giulia and Ugurel, Selma and Siveke, Jens and Gr{\"u}nwald, Barbara and Egger, Jan and others},
  journal={Medical Image Analysis},
  volume={94},
  pages={103143},
  year={2024},
  publisher={Elsevier}
}

@inproceedings{ruschhaupt_comparing_2025,
	title = {Comparing Mask R-{CNN} and Mask2Former architectures for individual tree crown delineation},
	doi = {10.18420/giljt2025_13},
    booktitle = {45. GIL-Jahrestagung, Digitale Infrastrukturen für eine nachhaltige Land-, Forst-und Ernährungswirtschaft},
	author = {Ruschhaupt, Sonja and Troles, Jonas and Schmid, Ute},
	date = {2025-02},
    year = {2025}
}

@article{lungo_vaschetti_treepseco_2025,
	title = {{TreePseCo}: Scaling Individual Tree Crown Segmentation using Large Vision Models},
	volume = {{XLVIII}-M-7-2025},
	url = {https://isprs-archives.copernicus.org/articles/XLVIII-M-7-2025/275/2025/},
	doi = {10.5194/isprs-archives-XLVIII-M-7-2025-275-2025},
	pages = {275--282},
	journal = {The International Archives of the Photogrammetry, Remote Sensing and Spatial Information Sciences},
	author = {Lungo Vaschetti, J. and Arnaudo, E. and Rossi, C.},
	year = {2025},
}

@misc{ben_weinstein_2022_5914554,
  author       = {Ben Weinstein and
                  Sergio Marconi and
                  Ethan White},
  title        = {Data for the NeonTreeEvaluation Benchmark},
  month        = jan,
  year         = 2022,
  publisher    = {Zenodo},
  version      = {0.2.2},
  doi          = {10.5281/zenodo.5914554},
  url          = {https://doi.org/10.5281/zenodo.5914554},
}

@misc{teng_assessing_2025,
      title={Assessing SAM for Tree Crown Instance Segmentation from Drone Imagery}, 
      author={Mélisande Teng and Arthur Ouaknine and Etienne Laliberté and Yoshua Bengio and David Rolnick and Hugo Larochelle},
      year={2025},
      eprint={2503.20199},
      archivePrefix={arXiv},
      primaryClass={cs.CV},
      url={https://arxiv.org/abs/2503.20199}, 
}

@article{tong_individual_2025,
	title = {Individual tree crown delineation in high resolution aerial {RGB} imagery using {StarDist}-based model},
	volume = {319},
	issn = {0034-4257},
	url = {https://www.sciencedirect.com/science/article/pii/S0034425725000227},
	doi = {https://doi.org/10.1016/j.rse.2025.114618},
	pages = {114618},
	journal = {Remote Sensing of Environment},
	author = {Tong, Fei and Zhang, Yun},
	year = {2025},
}

@inproceedings{veitch-michaelis_oam-tcd_2024,
author = {Veitch-Michaelis, Josh and Cottam, Andrew and Schweizer, Daniella and Broadbent, Eben N. and Dao, David and Zhang, Ce and Zambrano, Angelica Almeyda and Max, Simeon},
title = {OAM-TCD: a globally diverse dataset of high-resolution tree cover maps},
year = {2024},
isbn = {9798331314385},
publisher = {Curran Associates Inc.},
address = {Red Hook, NY, USA},
booktitle = {Proceedings of the 38th International Conference on Neural Information Processing Systems},
articleno = {1574},
numpages = {19},
location = {Vancouver, BC, Canada},
series = {NIPS '24}
}

@article{ball_accurate_2023,
  title={Accurate delineation of individual tree crowns in tropical forests from aerial RGB imagery using Mask R-CNN},
  author={Ball, James GC and Hickman, Sebastian HM and Jackson, Tobias D and Koay, Xian Jing and Hirst, James and Jay, William and Archer, Matthew and Aubry-Kientz, M{\'e}laine and Vincent, Gr{\'e}goire and Coomes, David A},
  journal={Remote Sensing in Ecology and Conservation},
  volume={9},
  number={5},
  pages={641--655},
  year={2023},
  publisher={Wiley Online Library}
}

@article{HE2022102667,
title = {Generating 2m fine-scale urban tree cover product over 34 metropolises in China based on deep context-aware sub-pixel mapping network},
journal = {International Journal of Applied Earth Observation and Geoinformation},
volume = {106},
pages = {102667},
year = {2022},
issn = {1569-8432},
doi = {https://doi.org/10.1016/j.jag.2021.102667},
url = {https://www.sciencedirect.com/science/article/pii/S0303243421003743},
author = {Da He and Qian Shi and Xiaoping Liu and Yanfei Zhong and Liangpei Zhang},
}

\end{document}